\documentclass[conference]{IEEEtran}
\IEEEoverridecommandlockouts
\usepackage{cite}
\usepackage{amsmath,amssymb,amsfonts}
\usepackage{algorithmic}
\usepackage{graphicx}
\usepackage{amsfonts}
\usepackage{textcomp}
\usepackage{bbm}
\usepackage{xcolor}
\usepackage{cite}

\usepackage{subcaption}
\usepackage{booktabs}
\usepackage{titlesec}

\titlespacing\section{0pt}{2pt plus 2pt minus 2pt}{2pt plus 0pt minus 2pt}
\titlespacing\subsection{0pt}{2pt plus 2pt minus 2pt}{2pt plus 0pt minus 2pt}
\titlespacing\subsubsection{0pt}{2pt plus 2pt minus 2pt}{2pt plus 0pt minus 2pt}

\def\BibTeX{{\rm B\kern-.05em{\sc i\kern-.025em b}\kern-.08em
    T\kern-.1667em\lower.7ex\hbox{E}\kern-.125emX}}
\begin{document}

\title{Clustering Egocentric Images in Passive Dietary Monitoring with Self-Supervised Learning\\
\thanks{This work is supported by the Innovative Passive Dietary Monitoring
Project funded by the Bill \& Melinda Gates Foundation (Opportunity ID:
INV-006713). $\dagger$ indicates the corresponding authors (jianing.qiu17@imperial.ac.uk and benny.lo@imperial.ac.uk)}
}

% \author{\IEEEauthorblockN{1\textsuperscript{st} Jiachuan Peng}
% \IEEEauthorblockA{\textit{Department of Surgery and Cancer} \\
% \textit{Imperial College London}\\
% London, UK \\
% jiachuan.peng21@imperial.ac.uk}
% \\
% \IEEEauthorblockN{4\textsuperscript{th} Xinwei Ju}
% \IEEEauthorblockA{\textit{Department of Surgery and Cancer} \\
% \textit{Imperial College London}\\
% London, UK \\
% xinwei.ju21@imperial.ac.uk}
% \and
% \IEEEauthorblockN{2\textsuperscript{nd} Peilun Shi}
% \IEEEauthorblockA{\textit{Department of Surgery and Cancer} \\
% \textit{Imperial College London}\\
% London, UK \\
% peilun.shi21@imperial.ac.uk}
% \\
% \IEEEauthorblockN{5\textsuperscript{th} Frank P.-W. Lo}
% \IEEEauthorblockA{\textit{Department of Surgery and Cancer} \\
% \textit{Imperial College London}\\
% London, UK \\
% po.lo15@imperial.ac.uk}
% \\
% \IEEEauthorblockN{7\textsuperscript{th} Benny Lo}
% \IEEEauthorblockA{\textit{Department of Surgery and Cancer} \\
% \textit{Imperial College London}\\
% London, UK \\
% benny.lo@imperial.ac.uk}
% \and
% \IEEEauthorblockN{3\textsuperscript{rd} Jianing Qiu}
% \IEEEauthorblockA{\textit{Department of Computing} \\
% \textit{Imperial College London}\\
% London, UK \\
% jianing.qiu17@imperial.ac.uk}
% \\
% \IEEEauthorblockN{6\textsuperscript{th} Xiao Gu}
% \IEEEauthorblockA{\textit{Department of Surgery and Cancer} \\
% \textit{Imperial College London}\\
% London, UK \\
% xiao.gu17@imperial.ac.uk}
% }

\author{Jiachuan Peng$^1$, Peilun Shi$^1$, Jianing Qiu$^{1, \dagger}$, Xinwei Ju$^1$, Frank P.-W. Lo$^1$, Xiao Gu$^1$, \\ Wenyan Jia$^2$, Tom Baranowski$^3$,
Matilda Steiner-Asiedu$^4$, Alex K. Anderson$^5$, Megan A McCrory$^6$, \\ Edward Sazonov$^7$, Mingui Sun$^2$, Gary Frost$^1$, and Benny Lo$^{1, \dagger}$\\
$^1$Imperial College London, $^2$University of Pittsburgh, $^3$Baylor College Of Medicine, $^4$University of Ghana, \\ $^5$University of Georgia, $^6$Boston University, $^7$University of Alabama\\

% {\tt\small \{jiachuan.peng21, p.shi21, jianing.qiu17, x.ju21, po.lo15, xiao.gu17, benny.lo\}@imperial.ac.uk}
% \and
}

\IEEEaftertitletext{\vspace{-2.0\baselineskip}}

\maketitle

\begin{abstract}
In our recent dietary assessment field studies on passive dietary monitoring in Ghana, we have collected over 250k in-the-wild images. The dataset is an ongoing effort to facilitate accurate measurement of individual food and nutrient intake in low and middle income countries with passive monitoring camera technologies. The current dataset involves 20 households (74 subjects) from both the rural and urban regions of Ghana, and two different types of wearable cameras were used in the studies. Once initiated, wearable cameras continuously capture subjects' activities, which yield massive amounts of data to be cleaned and annotated before analysis is conducted. To ease the data post-processing and annotation tasks, we propose a novel self-supervised learning framework to cluster the large volume of egocentric images into separate events. Each event consists of a sequence of temporally continuous and contextually similar images. By clustering images into separate events, annotators and dietitians can examine and analyze the data more efficiently and facilitate the subsequent dietary assessment processes. Validated on a held-out test set with ground truth labels, the proposed framework outperforms baselines in terms of clustering quality and classification accuracy.

\end{abstract}

\begin{IEEEkeywords}
Egocentric vision, self-supervised learning, clustering, multi-task learning, passive dietary assessment
\end{IEEEkeywords}

\section{Introduction}

The double burden of malnutrition~\cite{popkin2020dynamics} hinders the healthy growth and development of individuals in many low and middle income countries (LMICs). To reduce and eradicate malnutrition, a systematic understanding of the nutritional status and needs of the population living in LMICs is needed. The prerequisite for establishing this in-depth and systematic understanding is accurate dietary assessment. However, in nutritional epidemiology, conventional dietary assessment methods, such as 24-Hour dietary recall and food frequency questionnaire (FFQ) rely on subjective reports~\cite{shim2014dietary}, which are inaccurate and require a very labor-intensive process to collect and interpret the dietary data. In recent years, many technological approaches have emerged to assist or automate dietary assessment, and provide more accurate assessments. Existing technological approaches can be categorized as active or passive. Although dietary intake can be recorded in situ rather than retrospectively, active approaches still need user initiation and inputs, such as using the user's phone to take a picture of the meal before and after eating~\cite{naaman2021assessment}. This still introduces subjectivity and human bias, and only very limited information of dietary intake can be recorded. Passive approaches~\cite{qiu2020counting}, on the other hand, can record the entire episode of a subject eating a meal, without active participation from the subject. This enables the recorded data to be objective and the subject's eating routine to be minimally affected. Following the study protocol of our passive dietary monitoring~\cite{jobarteh2020development}, we used two different types of wearable cameras (eButton \cite{sun2015exploratory} and Automatic Ingestion Monitor \cite{farooq2016novel}) to capture, including but not limited to, dietary relevant activities, e.g., buying ingredients, processing food items, eating the meal, and resting, to provide a comprehensive recording of a subject's daily dietary intake and energy expenditure. This passive way of data recording is not limited to dietary monitoring. Many studies have also used wearable cameras to passively record egocentric videos/images of personal activities and social interactions~\cite{grauman2021ego4d} to conduct human-centric research.

\begin{figure}[!t]
\centerline{\includegraphics[width=\columnwidth]{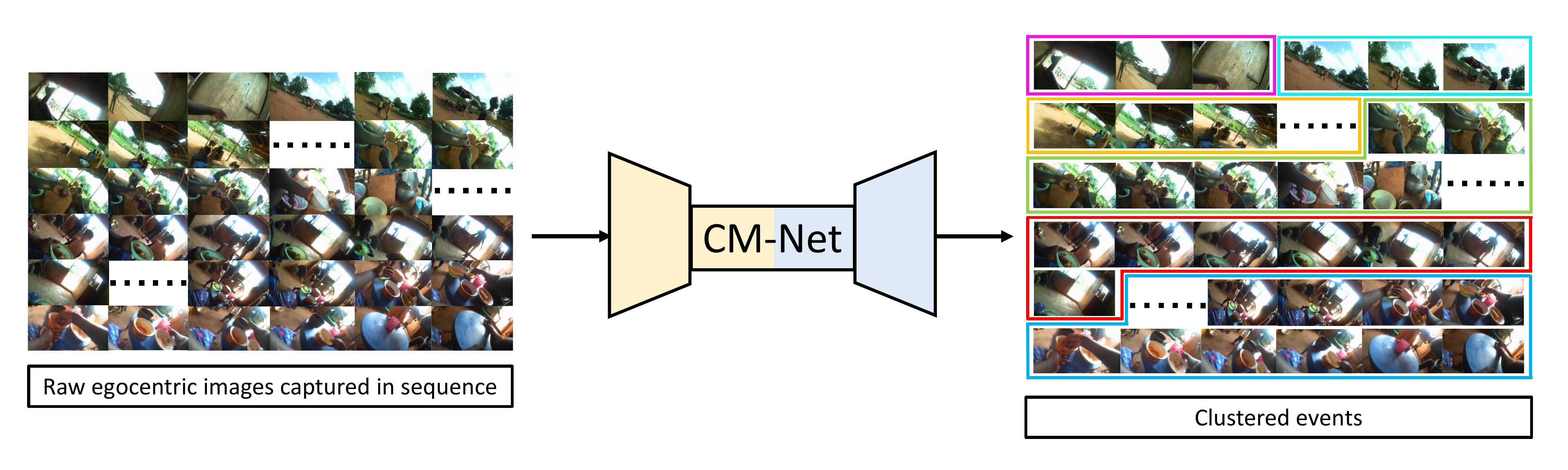}}
\caption{The proposed pipeline for clustering raw egocentric images into separate events. Each event (e.g., preparing a meal in the kitchen) is a sequence of temporally continuous and contextually similar images, in which the camera wearer performs a few or some actions and activities.}
\label{fig:raw2cluster_pipeline}
\end{figure}

However, continuously capturing data in a passive way generates a massive volume of unlabelled data. Manually trimming the long egocentric image sequence into separate events of interest and labelling them induce prohibitive costs. Therefore, in this work, we propose to cluster the massive egocentric images into separate events with a novel self-supervised learning framework, with the aim of simplifying and reducing the work load on data annotation and post-processing. After clustering, each cluster represents an independent event, and a long egocentric video/image stream is automatically separated into a set of events. The separate events can enable annotators and dietitians to understand the content of entire long video/image stream more easily, locate the events of interest more efficiently, and by focusing on each individual event, the subsequent annotation and analysis can be conducted at a more fine-grained level.

Thus, the contribution of our work is a novel self-supervised learning based framework to cluster massive volumes of egocentric images into separate events, which can simplify and accelerate the subsequent fine-grained data annotation and post-processing.

\section{Related Work}
\textbf{Image-based dietary assessment} can be divided into two types: active and passive. Specifically, active approaches~\cite{naaman2021assessment} require users to employ hand-held devices (e.g., smartphones) to capture images of food before and after the meal. Comparatively, passive approaches~\cite{jobarteh2020development, qiu2020counting} record an entire dietary intake episode automatically via wearable cameras, without active participation from the user. Recently, 360-degree cameras have also been proposed to passively record the dietary intake of all individuals in communal or food sharing scenarios~\cite{qiu2019assessing, lei2021assessing}. Although being more objective and comprehensive, passive approaches generate vast amounts of unlabelled data, as it continuously captures visual information, which leaves the subsequent post-processing and manual annotation prohibitively costly and time-consuming. Doulah and Sazonov~\cite{doulah2017clustering} investigated the use of clustering in dietary assessment, but their approach was histogram-based and focused on binary clustering (i.e., food and non-food images). Hand-crafted features were used to segment image sequences captured by wearable cameras~\cite{zhang2010segmentation}, whereas our method uses deep-based features learned with self supervision. 

\textbf{Self-supervised learning} aims to learn discriminative representations from data itself, without manual labelling, which is particularly suitable for processing massive volumes of egocentric images. Recent advances in self-supervised learning have shown that the representations learned in a self-supervised manner, when used in downstream tasks such as image recognition, can yield results comparable to those of supervised learning, which requires a large volume of ground truth labels for training. Contrastive learning is a type of self-supervised learning, the core of which is to distinguish similar and dissimilar data pairs. Chen et al.~\cite{chen2020simple} proposed SimCLR, a simple yet effective contrastive learning framework that exploits the composition of data augmentations as well as nonlinear transformation to learn discriminative image features in a self-supervised manner. Apart from contrastive learning, autoencoding has been widely used for self-supervised learning. Recently, masked autoencoder (MAE)~\cite{he2021masked} has been proposed, which combines autoencoding and pixel masking to learn discriminative image features.

\begin{figure}
\centering
\begin{subfigure}{.47\columnwidth}
  \centering
  \includegraphics[width=\columnwidth]{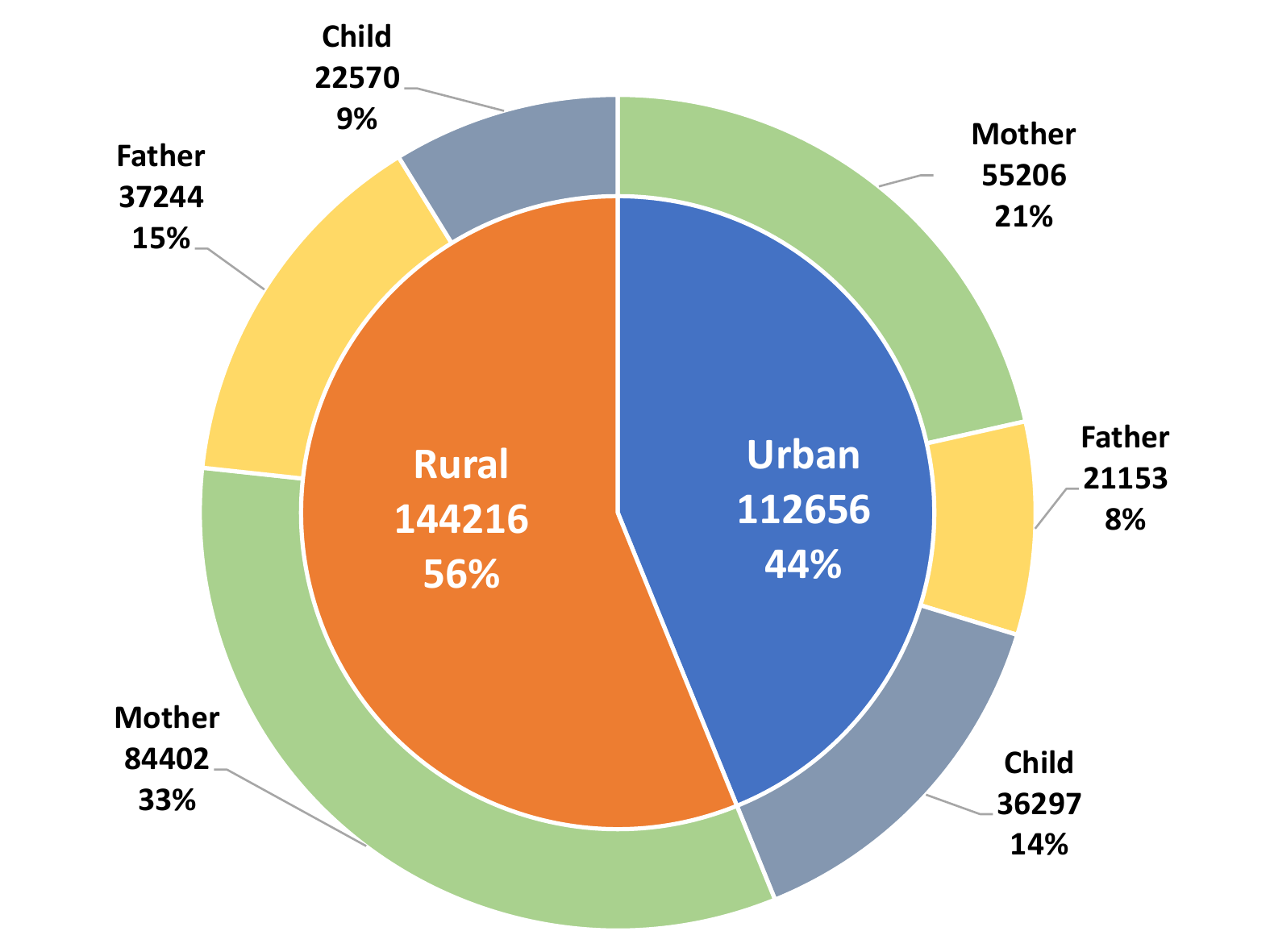}
  \caption{}
  \label{fig:dataset-L}
\end{subfigure}%
\begin{subfigure}{.53\columnwidth}
  \centering
  \includegraphics[width=\columnwidth]{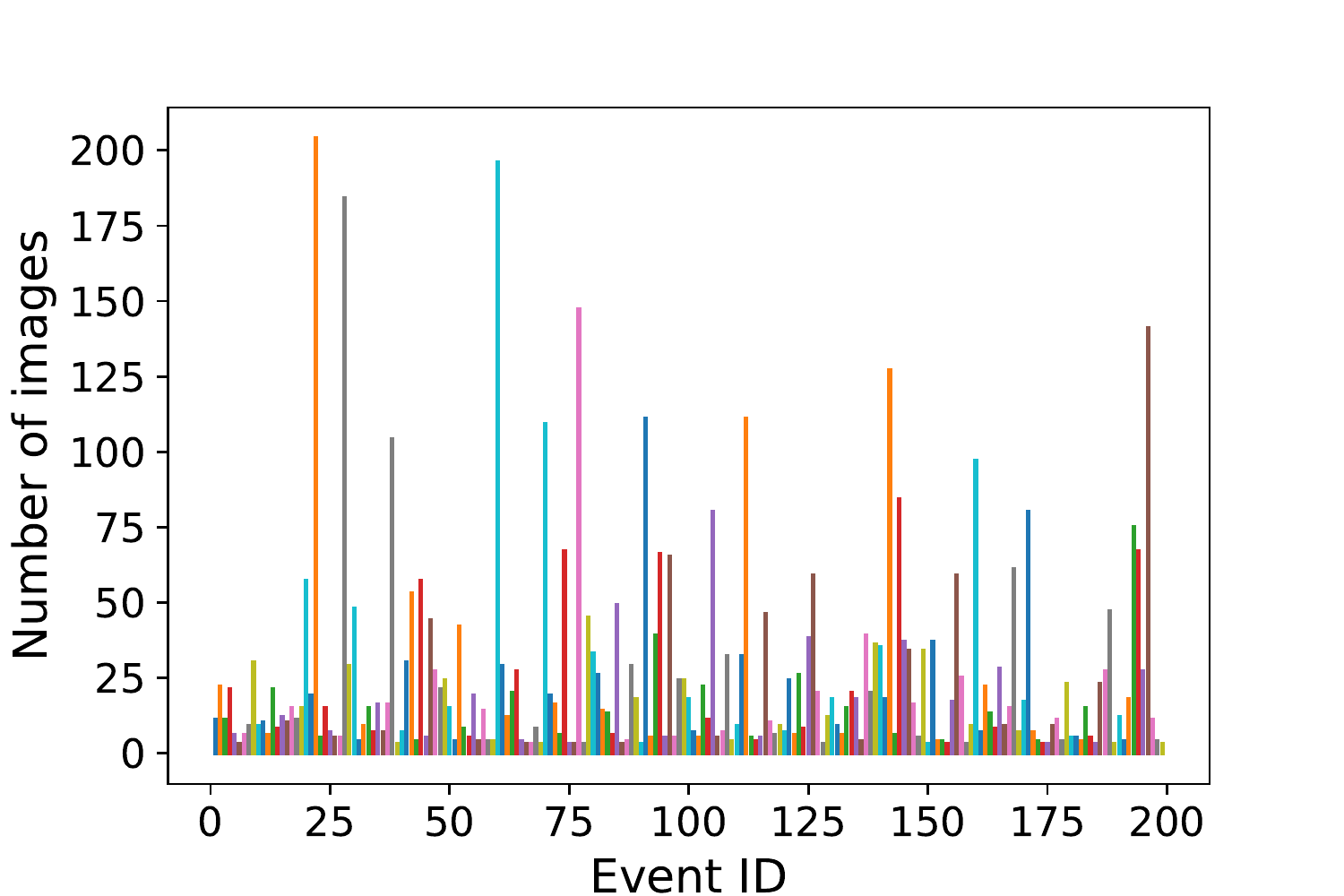}
  \caption{}
  \label{fig:dataset-S}
\end{subfigure}%
\caption{Statistics of Dataset-L (a) and Dataset-S (b)}
\label{fig:dataset_statistics}
\end{figure}

\section{Dataset}
The collected data was split into two sets: \textbf{Dataset-L} (Large) and \textbf{Dataset-S} (Small). Dataset-L contains 256,872 images, including different daily events captured with different individuals (e.g., mother, father, adolescent) from different households in rural and urban areas, but without manually annotated labels. Hence, this dataset is designed for pre-training the self-supervised learning framework. Detailed statistics of Dataset-L are shown in Fig.~\ref{fig:dataset-L}. Dataset-S has 4,954 images with 199 different dietary events. Similar to Dataset-L, those images were captured with different individuals from different households. Each image in Dataset-S is assigned a label indicating each event to which it belongs. Fig.~\ref{fig:dataset-S} shows the number of images each event has in Dataset-S. Dataset-S can be used to quantitatively evaluate the feature encoding capability of a self-supervised framework.

\section{Method}
\begin{figure}[!t]
\centerline{\includegraphics[width=\columnwidth]{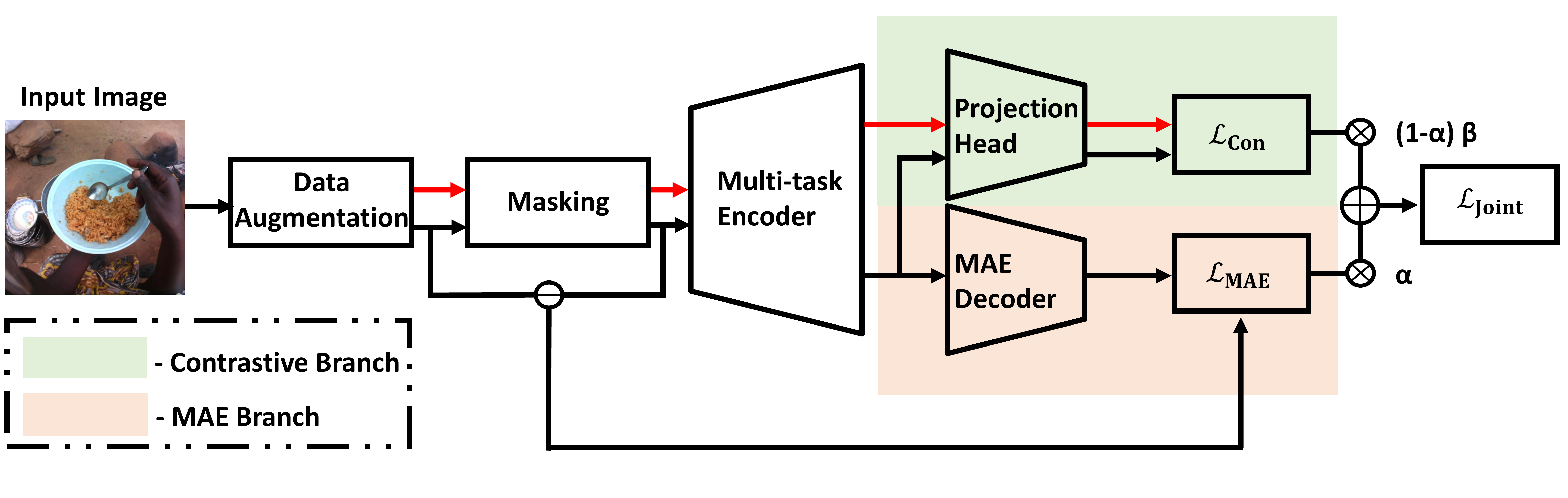}}
\caption{The proposed self-supervised learning framework (CM-Net) for learning discriminative egocentric image features. The framework consists of a multi-task encoder, and two parallel branches: a MAE branch and a contrastive learning branch. Two branches produce two losses,  which are combined to form a joint loss to update the framework's learnable parameters. After training the framework, we discard the projection head and MAE decoder, and encode images into representations with the multi-task encoder. The resulting representations are then used for clustering.}
\label{fig:framework}
\end{figure}

Self-supervised learning is used to cluster passively captured egocentric images into separate events. To this end, we designed a novel framework which uses a two-stream structure to combine MAE with contrastive learning. Fig.~\ref{fig:framework} shows our framework. The contrastive learning branch is devised to enable the encoder to learn subtle discrepancies between image frames. This is essential because egocentric images are captured consecutively, which means image frames within a certain time interval are similar and temporally correlated. The two branches are complementary and can enhance the overall learning capability as shown in Sec.~\ref{sec:results}

\subsection{Multi-task Encoder}\label{AA}
ViT-B\cite{dosovitskiy2020image} is used as the backbone encoder for multi-task learning as the two branches essentially perform two different tasks. We randomly apply the following data augmentation techniques to transform a single input image to two augmented views: resized cropping, flipping, color distortion, grayscale conversion, and Gaussian blurring. Before being fed into the encoder, some patches of augmented images are masked (masking rate is 75\%). Formally, given an input image X, we obtain two masked images $X_1,X_2=Mask(A_1,A_2)$; where $A_1,A_2=Aug(X)$ are the two augmented views of the input image $X$. Both $X_1$ and $X_2$ are then fed into the multi-task encoder (MTE) to generate their respective visual representations $h_1$ and $h_2$ as shown in Eq.~\ref{eq:mte}

\begin{equation}\label{eq:mte}
h_1,h_2=MTE(X_1,X_2)
\end{equation}

$h_1$ and $h_2$ can be extracted and used in downstream tasks once the entire framework has been properly trained.

\subsection{MAE Branch}
We adopted the MAE structure proposed in\cite{he2021masked}. Throughout the training, the MAE branch computes the mean squared error (MSE) on the masked patches between the reconstructed and original images as its loss. Eq.~\ref{eq:mae} shows the loss function.

\begin{equation}\label{eq:mae}
\mathcal{L}_{MAE}=MSE({Patch}_{recon},{Patch}_{orig})
\end{equation}
\begin{equation}
{Patch}_{recon} = {({MAE}_{decoder}(h_1+M_1+P_1))}_{masked}
\end{equation}
\begin{equation}
{Patch}_{orig} = A_1-X_1
\end{equation}
where $M_1$ and $P_1$ indicate the mask tokens and positional embedding, respectively.
\subsection{Contrastive Branch}
A projection head (PH), similar to the one in SimCLR~\cite{chen2020simple}, was designed and integrated into our framework following the multi-task encoder. We denote the output from the projection head as embeddings $z_1$ and $z_2$:
\begin{equation}
z_1,z_2=PH(h_1,h_2)\label{eq}
\end{equation}

Finally, the loss of the contrastive branch can be computed as shown in Eq.~\ref{eq:contrastive}:

\begin{equation}\label{eq:contrastive}
\resizebox{\columnwidth}{!}{
$\mathcal{L}_{Con}=-log(\frac{exp(Sim(z_{1,i},z_{2,i})/\tau)}{\sum_{j=1}^{N}{\mathbbm{1}}_{[j\ne i]}exp(Sim(z_{1,i},z_{1,j})/\tau) +  \sum_{k=1}^{N}exp(Sim(z_{1,i},z_{2,k})/\tau)})$}
\end{equation}

\begin{equation}\label{eq:sim}
Sim(z,z^\prime)=(\frac{\sum_{v=1}^W{cosine\_sim}(z_{C,v,H},z^\prime_{C,v,H})}{W})
\end{equation}
where $\tau$ is the temperature parameter and was set to 0.5 in our experiments. $\mathbbm{1}_{[j\ne i]}$ is 1 iff $j\ne i$ and is 0 otherwise. N is the batch size. The similarity function is shown in Eq.\ref{eq:sim}. Embeddings $z$ and $z^\prime$ are of shape (C,W,H), we compute the cosine similarity along the C and H axes, and subsequently average the results along the W axis. 

\subsection{Joint Loss}
In order to fully exploit the interaction between two branches, we use the loss of MAE branch as the main loss and that of contrastive branch as an auxiliary loss. As the common practice, contrastive learning benefits from a carefully-crafted data augmentation strategy. In light of this, for contrastive branch, on top of our chosen augmentations, MAE branch provides an additional augmentation, i.e., masking, which also vastly reduces the computational cost.

On the other hand, the knowledge learned in the contrastive branch, if combined properly, can be applied to the MAE branch. Thus, these two branches benefit from each other and result in better generalization and performance. The joint loss, which combines losses from both branches, is therefore designed as follows:
\begin{equation}
\mathcal{L}_{Joint}=\alpha \mathcal{L}_{MAE}+(1-\alpha)\beta \mathcal{L}_{Con}
\end{equation}
where \(\alpha\) is a weighting coefficient used to control how much each branch's loss contributes to the overall loss. \(\alpha\) was empirically set to 0.8 in our experiments. \(\beta\) is the scaling coefficient, with which we can scale the contrastive loss to the similar order of magnitude as MAE loss ($\beta$ = 0.02 in our experiments). We name our framework as CM-Net as it is based on contrastive learning and MAE.

\begin{table}[]
\centering
\caption{Linear Probing Results on Dataset-S}
\label{tab:lp-comparsion}
\begin{tabular}{@{}cc@{}}
\toprule
Method                & Accuracy (Top-1) \\ \midrule
A single MAE branch                   & 86.3\%           \\
A single masked constrastive learning branch   & 90.6\%           \\
A single unmasked constrastive learning branch & 92.2\%           \\ \midrule
CM-Net (Ours)         & \textbf{92.7\%}  \\ \bottomrule
\end{tabular}
\end{table}

\section{Results and Discussion}\label{sec:results}

We use the following methods as the baselines: 1) single MAE branch; 2) unmasked contrastive branch. It resembles an aforementioned conventional contrastive learning framework, which learns representations from two views of an image; and 3) masked contrastive branch. Compared to the unmasked branch, the input images are masked before being fed into the contrastive branch. All methods were trained with the AdamW optimizer 
till convergence (i.e., the loss ceases to evidently decrease). The learning rate was set to 5e-5, which reduced by a factor of 0.8 every 15 epochs. Batch size was 32 (16 when training the unmasked contrastive branch).

\subsection{Classification Accuracy of Linear Probing}

Linear probing has been a common practice to assess the quality of representations learned using self-supervised learning, which attaches a linear classifier after the encoder. Following this practice, we discarded the projection head and MAE decoder after self-supervised training the entire framework on Dataset-L. All weights of the encoder were then frozen and a linear classifier was added on top of the encoder. We then trained this linear classifier on Dataset-S (in a supervised manner). We only used random resized cropping and flipping to augment data, and trained the classifier for 50 epochs with AdamW. Table~\ref{tab:lp-comparsion} summarizes the best linear probing results of each method.

As shown in Table~\ref{tab:lp-comparsion}, our CM-Net achieves 92.7\% classification accuracy on Dataset-S. Comparatively, with a single MAE branch, the accuracy can only reach to 86.3\%. Note that masking degrades the classification accuracy of the contrastive branch (90.6\% vs. 92.2\%), which can be caused by the reduction of visual information. However, without masking, complete images are fed into the network. This increments both the computational cost and training time. In contrast, by combining the merits of MAE and contrastive learning, our framework reaches highest accuracy and halves the training time (48mins/epoch), compared to the unmasked contrastive branch (94mins/epoch). Training times were reported using a PC with an Intel i9 CPU and a NVIDIA 2080Ti GPU.

\subsection{Clustering Performance}
\begin{figure}[!t]
\centerline{\includegraphics[width=\columnwidth]{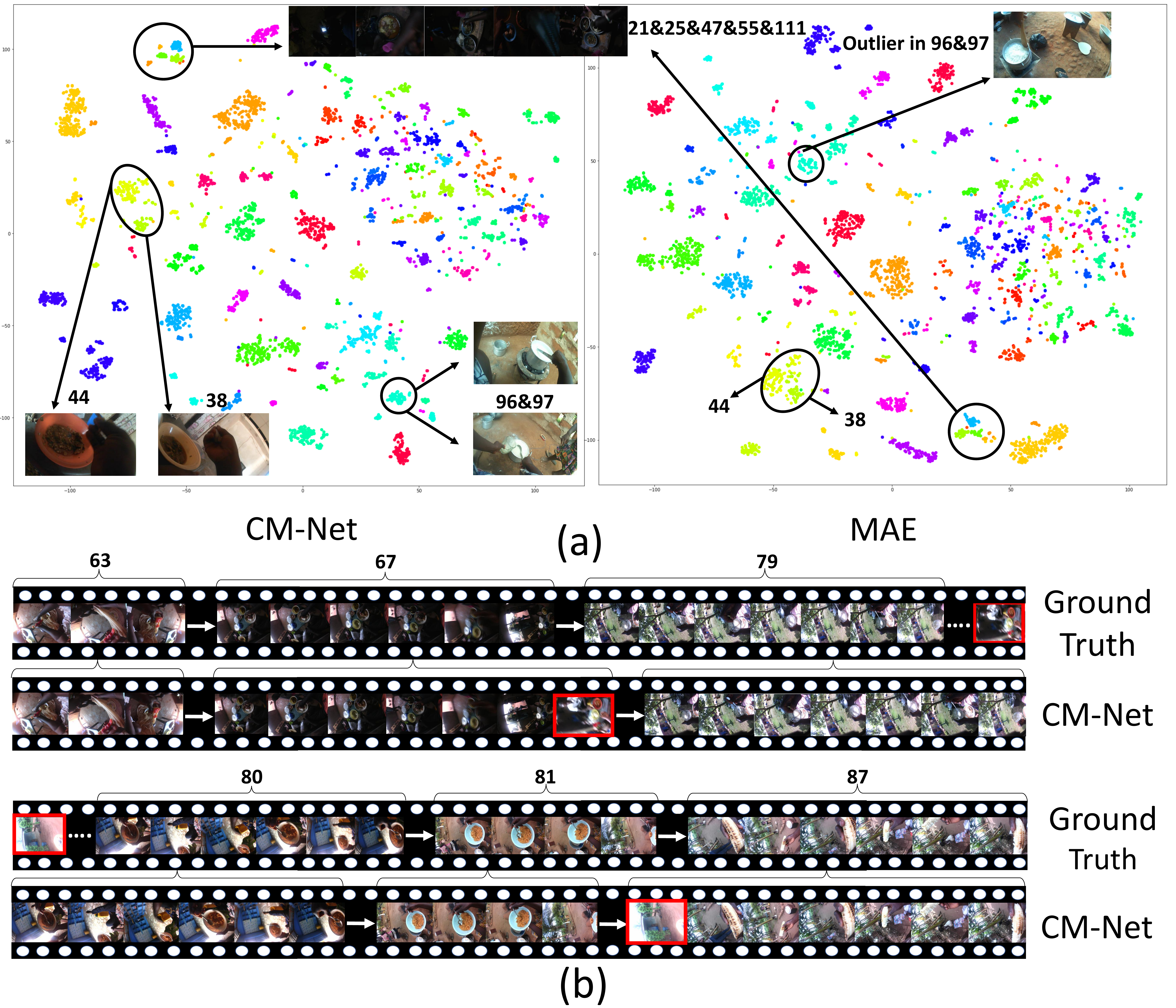}}
\caption{(a) t-SNE visualization of image representations encoded by our CM-Net and MAE. (b) Temporal comparison between the ground truth events and the events clustered by our CM-Net (red box indicates mis-aligned images). Results of (a) and (b) are from Dataset-S.}
\label{fig:cluster_seq}
\end{figure}

We use t-SNE\cite{van2008visualizing} to visualize feature clustering (each feature is associated with an image) as well as to examine the quality of the learned features by our multi-task encoder. Fig.~\ref{fig:cluster_seq}a shows the clustering results of our CM-Net and MAE on Dataset-S. Representative images for each cluster (indexes indicate different dietary events) are plotted on top.

As shown in Fig.~\ref{fig:cluster_seq}a, our CM-Net is able to merge images from the same event into a cluster where MAE fails, and better separate events with similar images than MAE. For example, event 44 resembles event 38 as they contain similar actions and objects. They both depict eating with a bowl (yellow and orange, respectively). CM-Net recognizes this difference and separates these two events distinctly whereas MAE clusters them together. Furthermore, even for alike dietary-related scenes at night (events 21\&25\&47\&55\&111), which contain little useful visual information, our CM-Net can still push them away from each other with certain distance. Additionally, successive events, e.g., 96\&97, can share the same context but with different activities conducted in steps. CM-Net recognizes that they are cooking and preparing banku, and separates them from other events distinctly while clustering them close. MAE in this case includes an outlier of cooking rice in events 96\&97. The visualized clusters verify that after self-supervised training, our framework's encoder can encode discriminative representations for sequentially captured egocentric images. Fig.\ref{fig:cluster_seq}b compares the events in temporal dimension between ground truth and those clustered by CM-Net.

\section{Conclusion}
We have proposed a novel self-supervised learning framework for clustering egocentric images into separate events. Evaluated with in-the-wild egocentric images, our proposed framework, which combines MAE and contrastive learning, was shown to be effective in learning discriminative egocentric image features in a self-supervised manner. By exploiting the masking strategy of the MAE branch, the contrastive learning branch improves the learned representations without much overhead added. Expanding the current dataset and validating our framework on other downstream tasks (e.g., image segmentation) are planned in future work.

\vspace{12pt}

\bibliographystyle{IEEEtran}

\bibliography{ref}
\end{document}